\documentclass[runningheads,a4paper]{llncs}
\usepackage{url}
\usepackage[lined,plain,commentsnumbered,linesnumbered]{algorithm2e}

\begin{document}
\mainmatter           

\title{Personality facets recognition from text}
\titlerunning{Personality facets recognition}


\author{
Wesley Ramos dos Santos \and
Ivandr\'e Paraboni 
}
\authorrunning{Santos and Paraboni} 
\tocauthor{Santos and Paraboni}

\institute{
	School of Arts, Sciences and Humanities, University of S\~ao Paulo\\
	Av. Arlindo Bettio, 1000. S\~ao Paulo, Brazil\\
	\email{\{wesley.ramos.santos,ivandre\}@usp.br}
}

\maketitle            

\begin{abstract}
Fundamental Big Five personality traits (e.g., Extraversion) and their facets (e.g., Activity) are known to correlate with a broad range of linguistic features and, accordingly, the recognition of  personality traits from text is a well-known Natural Language Processing task. Labelling text data with facets information, however, may require the use of lengthy personality inventories, and perhaps for that reason existing computational models of this kind are usually limited to the recognition of the fundamental traits. Based on these observations, this paper investigates the issue of personality facets recognition from text labelled only with information available  from a shorter personality inventory. In doing so, we provide a low-cost model for the recognition of certain personality facets, and present reference results for further studies in this field.
\keywords{Personality recognition, Big Five, facets}
\end{abstract}

\section{Introduction}
\label{sec-intro}

The Big Five personality model \cite{b5-goldberg} comprises five fundamental categories of personality - Extraversion, Agreeableness, Conscientiousness, Neuroticism, and Openness to experience - which are further divided into dozens of more specific {\em facets}. For instance, the Neuroticism  category includes facets representing  Anxiety, Depression etc. Big Five categories are strongly correlated to (and possibly defined by) language use and, as a result, the recognition of an individual's personality traits from text is a well-established task in the Natural Language Processing (NLP) field \cite{pan2015}.

Models for the recognition of personality traits from text are usually based on supervised machine learning methods that take as an input a text corpus labelled with personality scores. These scores, in turn, are computed from a range of  personality inventories (or questionnaires) such as the BFI-44 inventory \cite{b5-chapter}. The BFI-44 consists of a relatively short,  44 multiple-choice inventory conveying short items such as `I see myself as someone who {\em is depressed, blue}'. Items are to be answered on a zero (disagree strongly) to five (agree strongly) scale. 

Knowing the five fundamental categories of personality of an individual may be sufficient for a number of practical applications. For others, however, a more detailed assessment of personality facets may be called-for. Assessing personality facets usually involves the use of a more extensive personality inventory, such as the the 260-item NEO-PI-R \cite{neo}. From a computational perspective, however, large or complex inventories of this kind may be impractical, which may explain why studies on personality recognition from text \cite{mairesse-rec-jair,nowson-weblogs,pan2015,b5-propor} are usually limited to the five main personality categories obtainable from short inventories such as the BFI-44. 

Despite these difficulties, a compromise between convenience (as in the BFI-44) and expressiveness (as in NEO-PI-R) may still be possible. In particular, we notice that the work in \cite{facets} proved evidence that, although most facets cannot be explicitly captured by the BFI-44, a small subset of 10 facets (two from each of the main Big Five factors) are inferable from this short scale. Thus, it may be possible to obtain at least some of the facet labels available from NEO-PI-R at a much lower cost.

Based on these observations, the actual NLP question to be investigated in this paper is whether the 10 additional facets proposed in \cite{facets} may be automatically recognised from text labelled with BFI-44 information only. To this end, we developed a series of binary classifiers for Big Five facet recognition from a labelled  corpus of Brazilian Facebook status updates, and we present reference  results for further studies in this field. To the best of our knowledge, our work is the first attempt to learn personality facets in this way, and it is most likely the first of its kind to be devoted to the Brazilian Portuguese language.

\section{Related work}
\label{sec-background}

We are not aware of any large-scale work on Big Five facet recognition from text, but there is a wide range of studies focused on the more general task of recognising its main five personality categories. Given that the applicable methods are presumably similar, in what follows we briefly review a number of instances of the latter.

The work in \cite{mairesse-rec-jair} presents a comprehensive view of the personality recognition task from multiple computational perspectives (i.e., as classification, regression and ranking tasks), by comparing the use of written essays and speech corpus as input data, and by comparing the use of self-reported Big Five scores and those produced by specialists, among other issues. The study makes extensive use of psycholinguistic features provided by the LIWC \cite{liwc} and MRC \cite{mrc} databases, and results suggest that using ranking algorithms, speech as input data, and personality reports produced by specialists work best.

Contrary to the use of psycholinguistics-motivated features in \cite{mairesse-rec-jair} and others, the work in \cite{nowson-weblogs} makes use of n-gram models to classify extremes of  personality using both Naive-Bayes and SVM models. Evaluation based on a corpus of personal blogs achieves maximum accuracy of 65\%.

In the context of the PAN-CLEF shared task series \cite{pan2015}, a number of supervised models of personality recognition based on Twitter data labelled with personality scores obtained from a 10-item Big Five inventory have been developed. These include the overall winner of the competition \cite{pan2015-alvarez}, which combines second order attributes with a LSA text representation; the work in \cite{pan2015-gonzalez}, which makes use of char and POS n-gram models, and the work in \cite{pan2015-sulea}, which makes use of TF-IDF counts and stylistic features. For details, we refer to \cite{pan2015}.

\section{Personality facet recognition}
\label{sec-classes}

The present study aims to compare a number of models of personality facet recognition from text. More specifically, we consider the set of 10 personality facets that, according to the method discussed in \cite{facets}, may be inferred from the BFI-44 inventory \cite{b5-chapter} : Assertiveness and Activity facets (under the main Extraversion category), Altruism and Compliance (under Agreeableness), Order and Self-discipline (under Conscientiousness), Anxiety and Depression (under Neuroticism), and Aesthetics and Ideas (under Openness to experience.)  

The method proposed in \cite{facets} consists of a series of theoretically-motivated calculations (in addition to those already performed to obtain the basic Big Five personality scores) over the set of 44 responses provided by the BFI-44 inventory. Thus, provided that the full set of BFI-44 responses about an individual is known, computing these 10 additional facet scores is straightforward. 

For instance, according to \cite{facets}, the Activity facet of the Big Five Extraversion category is defined as the simple average of two of the  BFI-44 scores from which the main Extraversion score is obtained in the first place. In the present work, these facet scores are therefore taken as given, and we do not discuss the underlying method to obtain them. For details, see \cite{facets}.

Following existing work on Big Five personality recognition for the English language and others  \cite{mairesse-rec-jair,nowson-weblogs}, personality facet recognition is presently regarded as a set of independent binary classification tasks. To this end, a document is to be labelled as a positive instance of a given facet if the corresponding author shows an above-average score for that facet when considering the entire set of authors in the domain. Since personality facets are, by definition, independent from each other \cite{b5-goldberg}, each document is to be assigned ten individual labels corresponding to each facet, which are to be classified one at a time.

\section{Experiment}
\label{sec-exp}

\subsection{Overview}

We devised an experiment to compare three binary classifiers for personality facet recognition from text:

\begin{itemize}
	\item{BoW: bag-of-words features from the 3000 most frequent words in  corpus}
	\item{skip: average word vectors obtained from a skip-gram-1000 model}
	\item{cbow: average word vectors obtained from a cbow-1000 model}
\end{itemize}

The Bow model is built using Naive Bayes classification. Both skip and cbow models are built using logistic regression and pre-trained word embeddings computed from a 150-million Brazilian Twitter corpus using word2vec \cite{word2vec} with window size=5 and min\_count=10. In addition to these three classifiers, we  also consider a simple Majority class baseline system for illustration purposes.

\subsection{Data}

We use the 2.2 million-words {\em b5-post} corpus of Brazilian  Facebook \cite{b5-corpus}, conveying 194k status updates written by 1019 users, which are accompanied by self-reported BFI-44 \cite{b5-chapter} inventories filled-in by every user. 

The text portion of the corpus was subject to basic spell checking and term substitution (e.g., laugh expressions such as `haha' were replaced by a common \$LAUGH\$ symbol etc.)  From the corpus inventories, 10 additional personality facets were inferred according to the method in \cite{facets}. This information constitutes the set of ten class labels for each document as discussed in the previous section.

\subsection{Procedure}

All models were built using 10-fold cross validation over the entire {\em b5-post} dataset. However, since that we now intend to learn ten (facet) classes, and not only five (main categories), and since many facets may be considerably more sparse than others (e.g., the Depression facet of Neuroticism may be naturally less common than, say, Self-consciousness), data imbalance is a major concern to our work. As a means to alleviate this, we resort to  SMOTE  minority sampling with $k=5$ neighbours \cite{smote}.

\section{Results}
\label{sec-results}

Table \ref{tab-results} shows reference results for the majority class baseline, and for the three models of interest. The first column represents mean F1 scores over the ten classification tasks, followed by the number of times (wins) in which each model was the overall winner, and the mean F1 measure for each individual class.

\begin{table}
\caption{\label{tab-results}10-fold cross validation mean F1 scores for personality facets classification.}
\begin{center}
\begin{tabular}{l c c | c c c c c c c c c c}
\hline
model   & overall& wins& assert.& activ.& altr. & compl.& order & selfd.& ans.& depr.& aesth & ideas \\
\hline
Baseline& 0.33   & 0   & 0.33   & 0.34  & 0.34  & 0.34  & 0.33  & 0.33  &0.34 & 0.33 & 0.33  & 0.33 \\
   BoW  & 0.57   & 4   & 0.60   & 0.59  & 0.61  & 0.54  & 0.56  & 0.61  &0.57 & 0.52 & 0.60  & 0.58 \\
   skip & 0.58   & 4   & 0.60   & 0.58  & 0.62  & 0.52  & 0.55  & 0.62  &0.59 & 0.54 & 0.62  & 0.58 \\
   cbow & 0.58   & 7   & 0.60   & 0.59  & 0.60  & 0.54  & 0.55  & 0.60  &0.59 & 0.55 & 0.63  & 0.60 \\
\hline
\end{tabular}
\end{center}
\end{table}

Although all models present a considerable improvement over our admittedly simple baseline, the distinction among them is narrow, particularly between BoW and skip. A slight advantage of the cbow model over the others is however noticeable in the number of classes (wins) for which cbow was the overall winner (7 out of 10 classification tasks.) 

As it is usually the case in personality classification, some personality traits tend to be more evident from text than others. In the present setting, we notice that Compliance and Depression recognition were the most challenging tasks. However, it remains unclear whether these facets are less explicit in language use in general, or simply less explicit in our Facebook domain. 

Finally, we notice that the present results are generally similar to those observed in Big Five personality classification in English \cite{mairesse-rec-jair} and other languages, and also along the lines of previous studies on the recognition of the main Big Five categories from the {\em b5-post} corpus \cite{b5-mult,b5-ieee}.

\section{Final remarks}
\label{sec-discussion}

This paper presented a number of models of Big Five facet recognition from a Brazilian Portuguese Facebook corpus and corresponding BFI-44 information. Our study suggests that, not unlike basic Big Five categories, the ten facets proposed in \cite{facets} may be recognised from text with reasonable accuracy if compared to a simple baseline system. In other words, our experiments suggest that we may in principle develop supervised models of personality recognition at a level of abstraction more specific than those obtainable from existing work, and  without resorting to larger or more complex inventories to provide the required text labels.

The current work provides only initial reference  results for further studies in this field, and a number of possible improvements are left as future work. In particular, we envisage de use of larger word embedding models and alternative learning architectures for this task, and further evaluation work by directly comparing our results against text labelled with actual facet information.

\section{Acknowledgements}
This work received support by FAPESP grant \# 2017/06828-1 and  \mbox{2016/14223-0}.

\bibliography{refs}
\bibliographystyle{splncs04}

\end{document}